\documentclass{article}

    \PassOptionsToPackage{numbers, compress}{natbib}

\usepackage[final]{neurips_2019}

\usepackage[utf8]{inputenc} 
\usepackage[T1]{fontenc}    
\usepackage{hyperref}       
\usepackage{url}            
\usepackage{booktabs}       
\usepackage{amsfonts}       
\usepackage{nicefrac}       
\usepackage{microtype}      
\usepackage{amsmath}
\usepackage{symbols}
\usepackage{graphicx}
\usepackage{subcaption}
\usepackage{color}
\usepackage[normalem]{ulem}
\usepackage[dvipsnames]{xcolor}
\usepackage{algorithm}
\usepackage[noend]{algpseudocode}
\usepackage{transparent}
\usepackage[font=small]{caption}
\usepackage{multirow}
\newcommand*{\mybox}[1]{\framebox{#1}}

\newcommand{\rQ}{{\mathbf{q}}}
\newcommand{\rR}{{\mathbf{r}}}
\newcommand{\rO}{{\mathbf{o}}}
\newcommand{\rhatO}{{\mathbf{\hat{o}}}}
\newcommand{\rW}{{\mathbf{w}}}
\newcommand{\vcrpp}{{R2C{\scriptsize ++}}}
\newcommand{\shortmodel}{{TAB-VCR}}
\usepackage{listings}

\usepackage{tikz,siunitx}
\usetikzlibrary{shapes.geometric,shapes.symbols}

\newcommand{\tikzsymbol}[2][circle]{\tikz[baseline=-0.5ex]\node[inner
sep=2pt,shape=#1,draw,#2]{};}%
\newcommand{\freeze}[1][]{\tikzsymbol[rectangle]{fill=blue, minimum width=6pt, minimum height=5pt}}
\newcommand{\tune}[1][]{\tikzsymbol[rectangle]{fill=orange, minimum width=6pt, minimum height=5pt}}
\newcommand{\present}[1][]{\tikzsymbol[rectangle]{fill=green, minimum width=6pt, minimum height=5pt}}
\newcommand{\absent}[1][]{\tikzsymbol[rectangle]{fill=red, minimum width=6pt, minimum height=5pt}}

\title{
TAB-VCR: Tags and Attributes based Visual Commonsense Reasoning Baselines
}

\author{
  Jingxiang Lin, Unnat Jain, Alexander G. Schwing
    \\
  University of Illinois at Urbana-Champaign\\
  \url{https://deanplayerljx.github.io/tabvcr} \\
}

\begin{document}

\maketitle

\begin{abstract}
Reasoning is an important ability that we learn from a very early age. Yet, reasoning is extremely hard for algorithms. Despite impressive recent progress that has been reported on tasks that necessitate reasoning, such as visual question answering and visual dialog, models often exploit biases in datasets. To develop models with better reasoning abilities, recently, the new visual commonsense reasoning (VCR) task has been introduced. Not only do models have to answer questions, but also do they have to provide a reason for the given answer. The proposed baseline achieved compelling results, leveraging a meticulously designed model composed of LSTM modules and attention nets. Here we show
that a much simpler  model obtained by ablating and pruning the existing intricate baseline can perform better with half the number of trainable parameters. By associating visual features with attribute information and better text to image grounding, we obtain further improvements for our simpler \& effective baseline, \textbf{\shortmodel}.
We show that this approach results in 
a 
5.3\%, 4.4\% and 6.5\% absolute improvement over the previous state-of-the-art~\cite{zellers2019vcr} on  question answering, answer justification and holistic VCR.

\end{abstract}
\section{Introduction}

Reasoning abilities are important for many tasks such as answering of (referential) questions, discussion of concerns and participation in debates. While we are trained to ask and answer ``why'' questions from an early age and while we generally master  answering of  questions about observations with ease, 
visual reasoning abilities are all but simple for algorithms. 

Nevertheless, respectable accuracies have been achieved recently for many tasks where visual reasoning abilities are necessary. For instance, for visual question answering~\cite{AnatolICCV2015,Goyal2017MakingTV} and visual dialog~\cite{visdial}, compelling results have been reported in recent years, and many present-day models achieve accuracies well beyond random guessing on challenging datasets such as~\cite{GaoNIPS2015,krishna2017visual,ZhuCVPR2016,Hudson2019GQAAN}. However, it is also known that algorithm results are not stable at all and trained models often leverage biases to answer questions. For example, both questions about the existence and non-existence of a ``pink elephant'' are likely answered affirmatively, while questions about counting are most likely answered with the number 2.  Even more importantly, a random answer is returned if the model is asked to explain the reason for the provided answer.

To address this concern,  a new challenge on ``visual commonsense reasoning''~\cite{zellers2019vcr} was introduced recently, combining reasoning about physics~\cite{MottaghiECCV2016,YeECCV2018}, social interactions~\cite{AlahiCVPR2016,VicolCVPR2018,ChuangCVPR2018,GuptaCVPR2018},   understanding of procedures~\cite{ZhouAAAI2018,AlayracCVPR2016} and forecasting of actions in videos~\cite{SinghWACV2016,EhsaniCVPR2018,zhou2015temporal,VondrickCVPR2016,FelsenCVPR2017,RhinehartICCV2017,YoshikawaARXIV2018}. In addition to answering a question about a given image, the algorithm is tasked to provide a rationale to justify the given answer. In this new dataset the questions, answers, and rationales are expressed using a natural language containing references to the objects. The proposed model, which achieves compelling results, leverages those cues by combining a long-short-term-memory (LSTM) module based deep net with attention over objects to obtain grounding and context. 

\begin{figure}[t]
    \centering
    \setlength{\tabcolsep}{0pt}
    \begin{tabular}{@{\hskip-0pt}c@{\hskip-0pt}c}
    \includegraphics[width=0.45\textwidth]{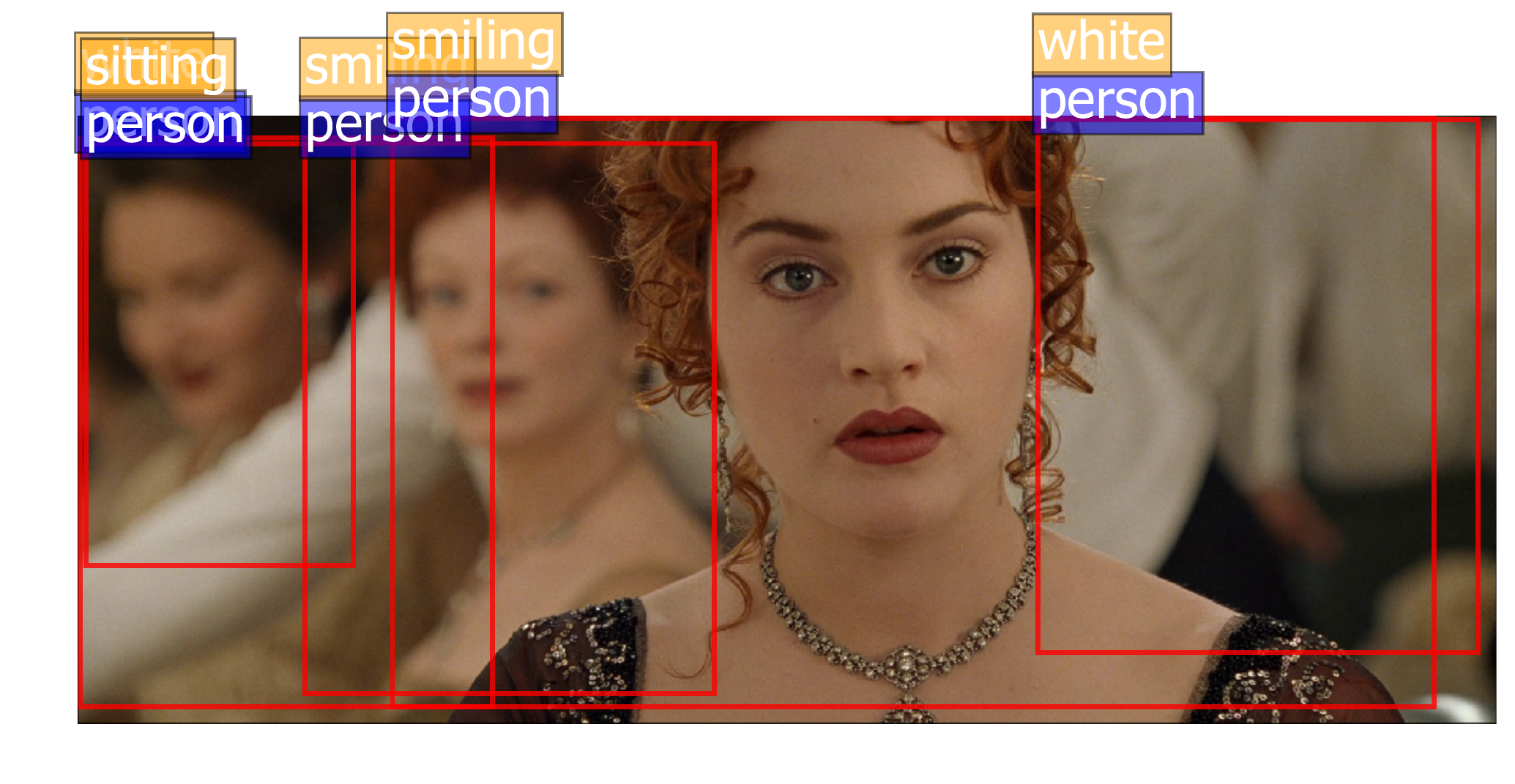} &
    \includegraphics[width=0.45\textwidth]{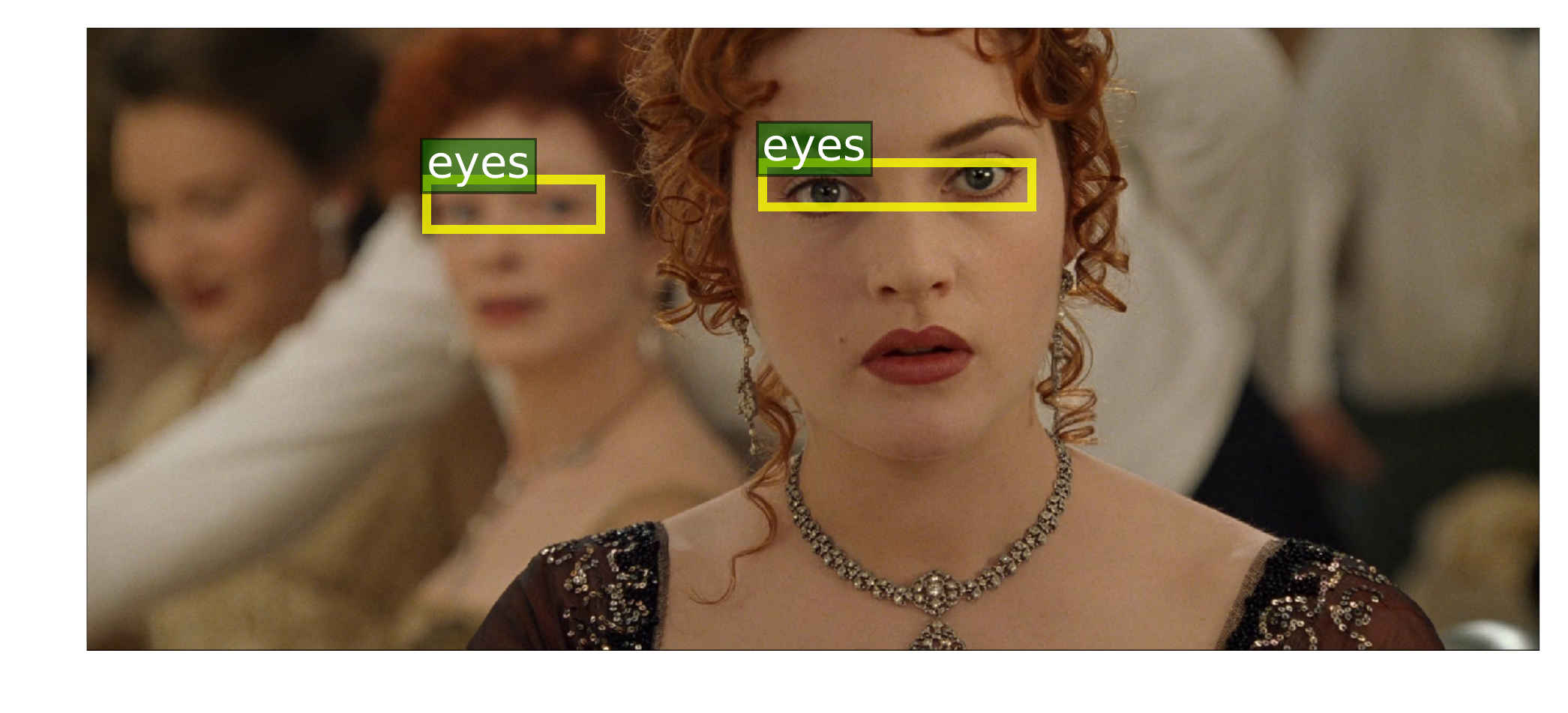}\\[-0.3cm]
    {\small (a) Associating attributes to VCR tags}&
    {\small (b) Finding tags missed by VCR}\\
    \end{tabular}
    \vspace{-0.3cm}
    \caption{\textbf{Motivation and improvements}. (a) The VCR object detections, \ie, red boxes and labels in \colorbox{Violet}{\color{white}blue} are shown. 
    We capture visual attributes by replacing the image classification CNN (used in previous models) with an image+attribute classification CNN. The predictions of this CNN are highlighted in \colorbox{Dandelion!70}{\color{white}orange}.
    (b) Additionally, many nouns referred to in the VCR text aren't \emph{tagged}, \ie grounded to objects in the image. We utilize the same image CNN as (a) to detect objects and ground them to text. The \emph{new tags} we found augment the VCR tags, and are highlighted with yellow bounding boxes and the associated labels in \colorbox{OliveGreen!70}{\color{white}green}. } 
    \label{fig:idea23}
     \vspace{-0.5cm}
\end{figure}

However, the proposed model is also very intricate. In this paper we revisit this baseline and show that a much simpler model with less than half the trainable parameters achieves significantly better results. As illustrated in~\figref{fig:idea23}, different from existing models, we also show that attribute information about objects and careful detection of objects can greatly improve the model performance.
To this end we extract visual features using an image CNN trained for the auxillary task of attribute prediction. In addition to encoding the image, we utilize the CNN to augment the object-word groundings provided in the VCR dataset. An effective grounding for these \emph{new tags} is obtained by using a combination of part-of-speech tagging and Wu Palmer similarity. We refer to our developed tagging and attribute baseline as \textbf{\shortmodel}.

We evaluate the proposed approach on the challenging and recently introduced visual commonsense reasoning (VCR) dataset~\cite{zellers2019vcr}. We show that a simple baseline which carefully leverages attribute information and object detections is able to outperform the existing state-of-the-art by a large margin despite having less than half the trainable model parameters. 

\vspace{-0.2cm}
\section{Related work}
\vspace{-0.2cm}
In the following we briefly discuss work related to vision based question answering, explainability and visual attributes.

\textbf{Visual Question Answering.} Image based question answering has continuously evolved in recent years, particularly also due to the release of various datasets~\cite{MalinowskiNIPS2014,RenNIPS2015,AnatolICCV2015,yu2015Madlibs,GaoNIPS2015,zhang2016yin,ZhuCVPR2016,krishna2017visual,JohnsonCVPR2017Clevr}. Specifically, \citet{zhang2016yin} and \citet{Goyal2017MakingTV} focus on balancing the language priors of \citet{AnatolICCV2015} for abstract and real images. \citet{agrawal2018don} take away  the IID assumption to create different distributions of answers for train and test splits, which further discourages transfer of language priors. \citet{Hudson2019GQAAN} balance open questions in addition to binary questions (as in \citet{Goyal2017MakingTV}). Image based dialog \cite{visdial,Vries2017GuessWhatVO,visdial_rl,JainCVPR2018,LuNIPS2017BestBothWorlds} can also be posed as a step by step image based question answering and question generation~\cite{mostafazadeh2016generating,JainCVPR2017,li2018visual} problem. Similarly related are question answering datasets built on videos \cite{tapaswi2016movieqa,maharaj2017dataset,lei2018tvqa,lei2019tvqa} and those based on visual embodied agents \cite{GordonCVPR2018,DasCVPR2018}. 

Various models have been proposed for these tasks, particularly for VQA~\cite{AnatolICCV2015,Goyal2017MakingTV}, selecting sub-regions of an image~\cite{tommasi2019combining}, single attention~\cite{chen2015abc,YangCVPR2016,AndreasCVPR2016,DasARXIV2016,FukuiARXIV2016,ShihCVPR2016,XuARXIV2015,ilievski2016focused,Yu2017MultimodalFB}, multimodal attention~\cite{LuARXIV2016,SchwartzNIPS2017,nam2017dual}, memory nets and knowledge bases~\cite{XiongICML2016,wu2016ask,wang2017explicit,ma2018visual}, improvements in neural architecture~\cite{MalinowskiICCV2015, MaARXIV2015,andreas2016learning,
andreas2016neural} and bilinear pooling representations~\cite{FukuiARXIV2016,kim2017hadamard,BenyounesICCV2017Mutan}.

\textbf{Explainability.} The effect of explanations on learning have been well studied in Cognitive Science and Psychology~\cite{lombrozo2012explanation,
williams2010role,
williams2013explanation}. Explanations play a critical role in child development~\cite{legare2014selective,crowley1999explanation} and more generally in educational environments~\cite{chi1989self,roscoe2008tutor,ross1995giving}. Explanation based models for applications in medicine \& tutoring have been previously proposed~\cite{shortliffe1975model,Lent2004AnEA,
Lane2005ExplainableAI,
core2006building}. Inspired by these findings, language and vision research on attention mechanism help to provide insights into decisions made by deep net models~\cite{LuARXIV2016,selvaraju2017gradcam}. Moreover, explainability in deep models has been investigated by modifying CNNs to focus on object parts~\cite{zhang2019interpreting,
zhang2018interpretable}, decomposing questions using neural modular substructures~\cite{andreas2016neural,andreas2016learning,DasECCV2018}, and interpretable hidden units in deep models~\cite{netdissect2017,bau2019gandissect}. Most relevant to 
our research
are works on natural language explanations. This includes multimodal explanation~\cite{park2018multimodalexp} and textual explanations for classifier decisions~\cite{hendricks2016generating} and self driving vehicles~\cite{Kim2018ECCV}.  

\begin{figure}[t]
\vspace{-0.1cm}
  \centering
  \begin{tabular}{@{\hskip-10pt}c@{\hskip-13pt}c}
  \includegraphics[width=0.3\linewidth,clip,trim={0 0.5cm 0 0}]{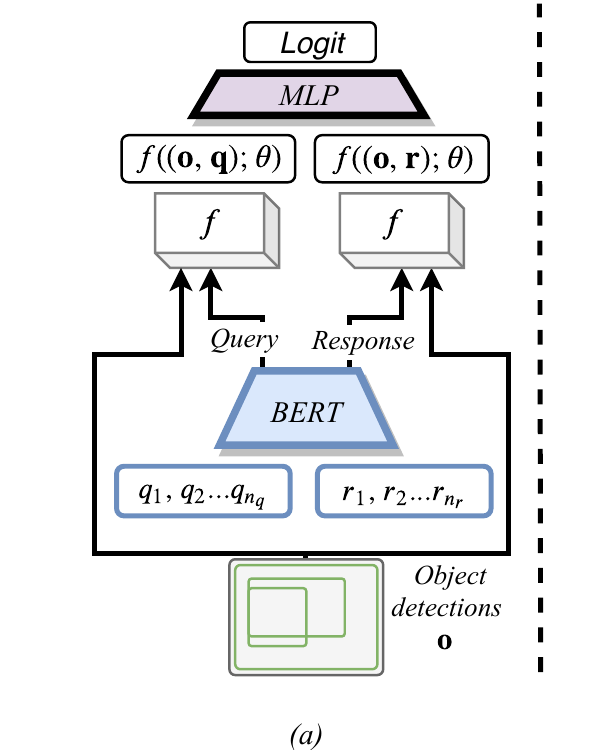}&
  \includegraphics[width=0.77\linewidth,clip,trim={0 0.5cm 0 0}]{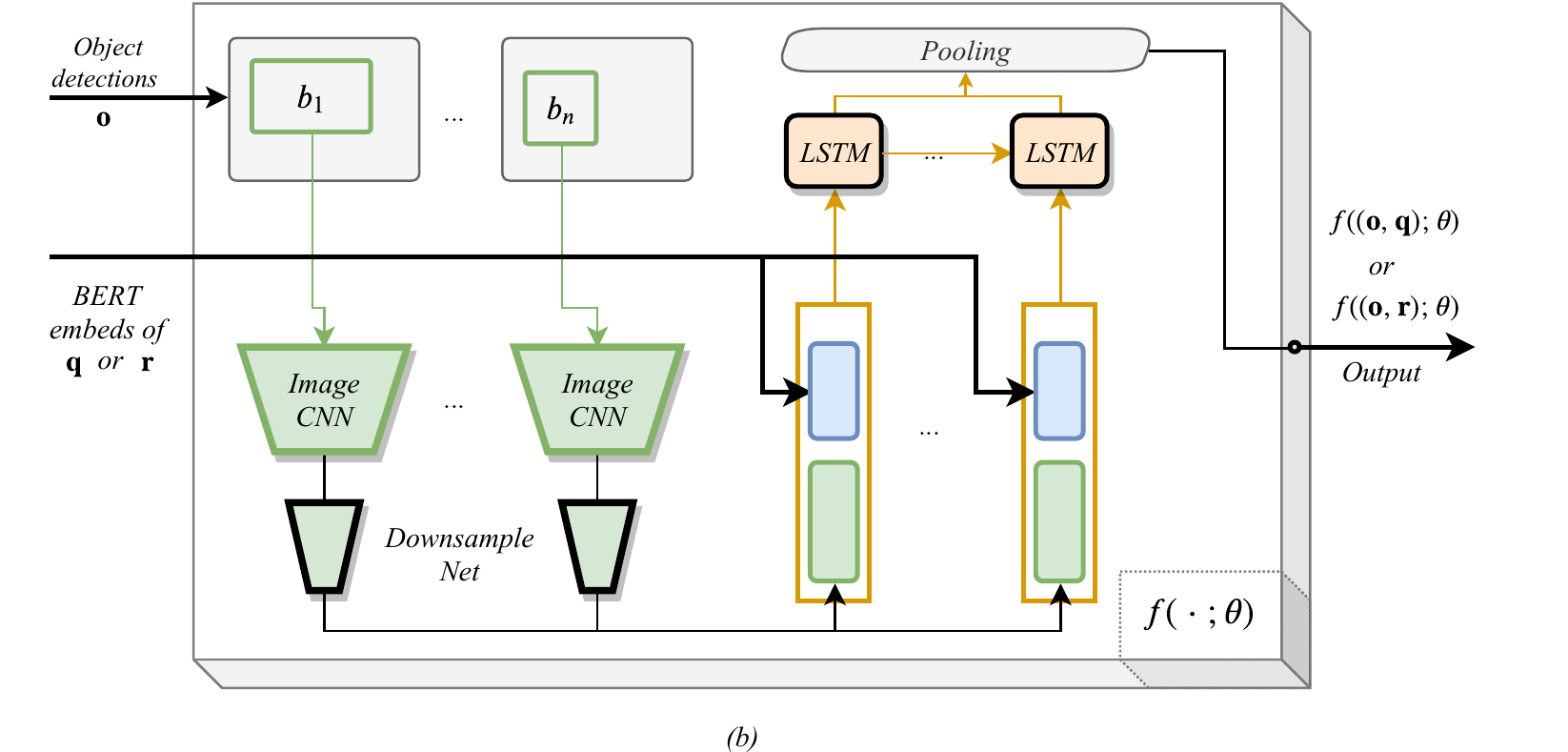}\\
  {\small (a) Overview}&
  {\small (b) Joint image \& language encoder}\\
  \end{tabular}
  \vspace{-0.2cm}
  \caption{(a) \textbf{Overview of the proposed~\shortmodel~model}: Inputs are the image (with object bounding boxes), a query and a candidate response. Sentences (query \& response) are represented using BERT embeddings and encoded jointly with the image using a deep net module $f(\cdot ; \theta)$. The representations of query and response are concatenated and scored via a multi-layer perceptron (MLP); (b) \textbf{Details of joint image \& language encoder} $f(\cdot ; \theta)$:
  BERT embeddings of each word are concatenated with their corresponding local image representation.
  This information is pass through an LSTM and pooled to give the output $f((I,\mathbf{w});\theta)$. The network components \mybox{outlined in black}, \ie, MLP, downsample net and LSTM are the only components with trainable parameters.}
  \label{fig:model}
  \vspace{-0.4cm}
\end{figure}

\textbf{Visual Commonsense Reasoning.} The recently introduced Visual Commonsense Reasoning dataset~\cite{zellers2019vcr} combines the above two research areas,  studying explainability (reasoning) through two multiple-choice  subtasks. First,  the question answering subtask requires to predict the answer to a challenging question given an image. Second, and more connected to explainability, is the answer justification subtask, which requires to predict the rationale given a question and a correct answer. To solve the VCR task,~\citet{zellers2019vcr} base their model on a convolutional neural network (CNN) trained for classification. Instead, we associate VCR detections with visual attribute information to obtain significant improvements with no architectural change or additional parameter cost. We discuss related work on visual attributes in the following. 

\textbf{Visual attributes.} Attributes are semantic properties to describe a localized object. Visual attributes are helpful to describe an unfamiliar object category~\cite{farhadi2009describing,lampert2009learning,russakovsky2010attribute}. Visual Genome~\cite{krishna2017visual} provides over 100k images along with their scene graphs and attributes. \citet{anderson2018bottom} capture attributes in visual features by using an auxiliary attribute prediction task on a ResNet101~\cite{he2016deep} backbone.

\vspace{-0.2cm}
\section{Attribute-based Visual Commonsense Reasoning}
\label{sec:approach}
\vspace{-0.2cm}
We are interested in visual commonsense reasoning (VCR). Specifically, we  study simple yet effective models and incorporate 
important information missed by previous methods -- attributes and additional object-text groundings.
Given an input  image, the VCR task is divided into two subtasks: (1) \textbf{question answering} (\qtoa): 
given a question (Q), select the correct answer (A)   from four candidate answers; (2) \textbf{answer justification} (\qator): given a question (Q) and its correct answer (A), select the correct rationale (R) from four candidate rationales. Importantly, both subtasks can be unified: choosing a \emph{response} from four options given a  \emph{query}. For \qtoa, the query is a question and the options are candidate answers. For \qator, the query is a question appended by its correct answer and the options are candidate rationales. Note, the \qtoar~task combines both, \ie, a model needs to succeed at both  \qtoa~and \qator. The proposed method focuses on choosing a response given a query, for which we   introduce  notation next. 

We are given an \emph{image}, a \emph{query}, and four candidate \emph{responses}. The words in the query and responses are grounded to objects in the image. The query and response are collections of words, while the image data is a collection of object detections. One of the detections also corresponds to the entire image, symbolizing a global representation. The image data is denoted by the set $\rO = (o_{i})_{i=1}^{n_{o}}$, where each $o_i$, $i\in \{1, \ldots, n_o\}$, consists of a bounding  box $b_i$ and a class label $l_i \in \cal{L}$\footnote{The dataset also includes information about segmentation masks, which are neither used here nor by previous methods. Data available at: \url{visualcommonsense.com}}. The query is composed of a sequence $ \rQ = (q_{i})_{i=1}^{n_q}$, where each $q_i$, $i\in\{1, \ldots, n_q\}$, is either a word in the vocabulary $\cal{V}$ or  a tag referring to a bounding box in $\rO$. A data point consists of four responses and we denote a response by the sequence $\rR = (r_{i})_{i=1}^{n_r}$, where $r_i$, $i\in\{1, \ldots, n_r\}$, (like the query) can either refer to a word in the vocabulary $\cal{V}$ or  a tag. 

We develop a conceptually simple joint encoder for language and image information, $f(\hspace{2pt}\cdot\hspace{2pt}; \theta)$, where $\theta$ is the catch-all for all the trainable parameters.

In the remainder of this section, we first present an overview of our approach. Subsequently, we discuss details of the joint encoder $f(\hspace{2pt}\cdot\hspace{2pt}; \theta)$. Afterward, we introduce 
how to incorporate attribute information and find \emph{new tags}, 
which helps improve the performance of our simple baseline. 
We defer details about training and implementation to the supplementary material. 

\vspace{-0.2cm}
\subsection{Overview}
\label{sec:overview}
\vspace{-0.2cm}
As mentioned, visual commonsense reasoning  requires to choose a response from four candidates. Here, we score each candidate separately. The separate scoring of responses is necessary to build a more widely applicable framework, which is independent of the number of responses to be scored. 

\begin{figure}
    \centering
    
    \setlength{\tabcolsep}{0pt}
    \begin{tabular}{@{\hskip-0pt}c}
    \includegraphics[width=0.9\textwidth]{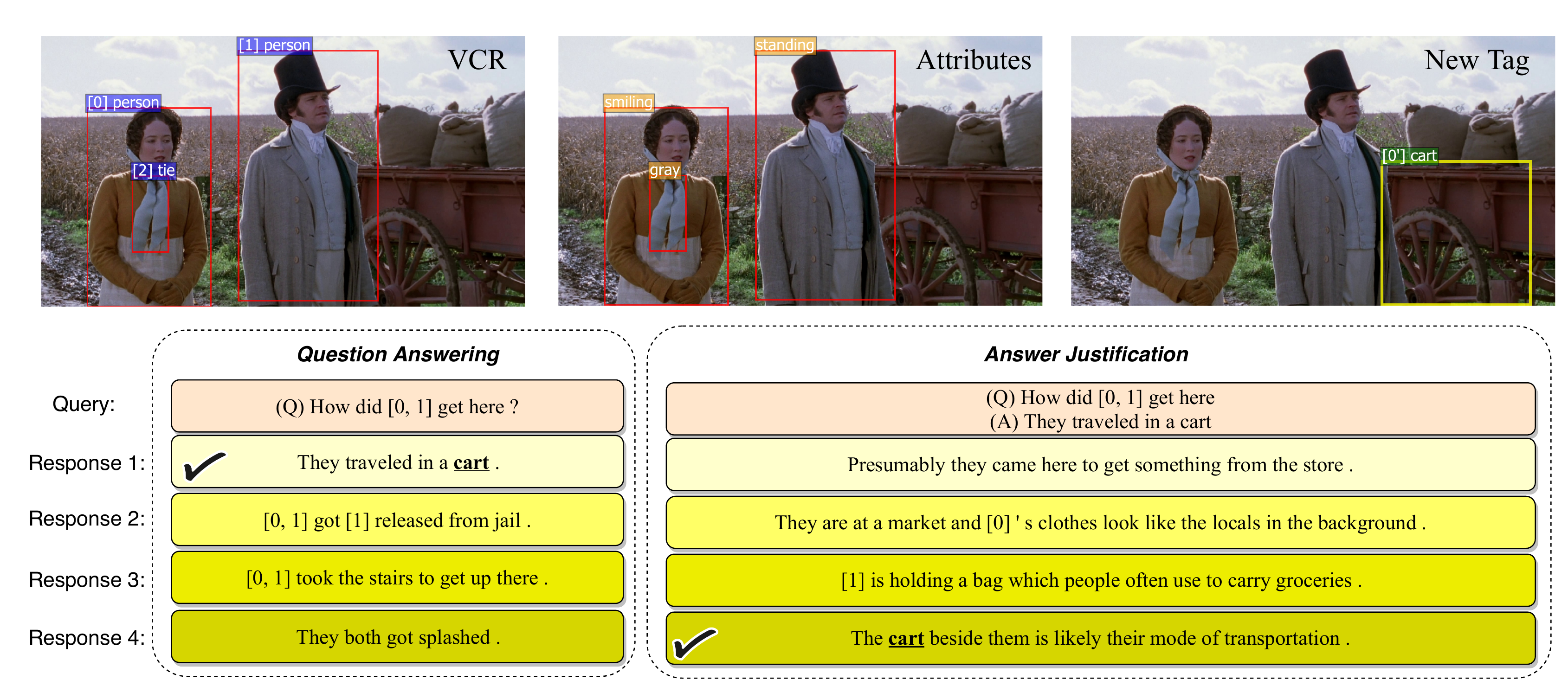} \vspace{-2mm}\\
    \small (a) Direct match of word \textbf{\uline{cart}} (in text) and the same label (in image).\\
    \includegraphics[width=0.9\textwidth]{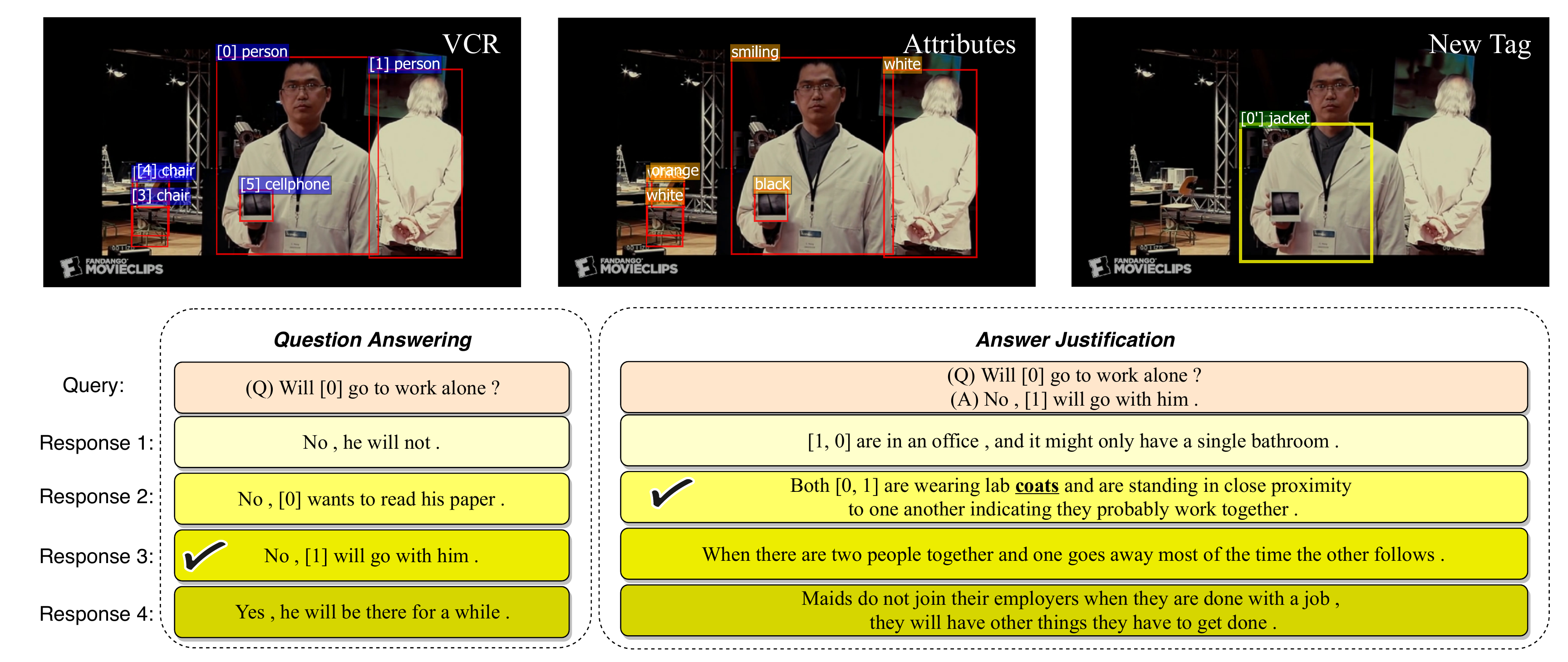}\vspace{-2mm}\\
    \small (b) Word sense based match of word \textbf{\uline{coats}} and label `jacket' with the same meaning.\vspace{-2mm}\\
    \end{tabular}
    \caption{\textbf{Qualitative results:} Two types of \emph{new tags} found by our method are (a) direct matches and (b) word sense based matches. Note that the images on the left show the object detections provided by VCR. The images in the middle show the attributes predicted by our model and thereby captured in visual features. The images on the right show \emph{new tags} detected by our proposed method. Below the images are the question answering and answer justification subtasks. }
    \label{fig:qual_results}
    \vspace{-0.5cm}
\end{figure}

Our proposed approach is outlined in \figref{fig:model}(a). 
 The three major components of our approach are: (1) BERT~\cite{devlin2018bert} embeddings for words; (2) a joint encoder $f(\hspace{2pt}\cdot\hspace{2pt}; \theta)$ to obtain $(\rO,\rQ)$ and $(\rO,\rR)$ representations; and (3) a multi-layer perceptron (MLP) to score these representations. 
 Each word in the query set $\rQ$ and response set $\rR$ is embedded via BERT. The BERT embeddings of $\rQ$ and associated image data from $\rO$ are jointly encoded to obtain the representation $f((\rO,\rQ);\theta)$. 
 An analogous representation for responses is obtained via $f((\rO,\rR);\theta)$. Note that the joint encoder is identical for both the query and the response. The two representations are concatenated and scored via an MLP. These scores or logits are further normalized using a \texttt{softmax}. The network is trained end-to-end using a cross-entropy loss of predicted probabilities vis-\`a-vis correct responses.
 
 Next, we provide details of the joint encoder before we describe our approach to incorporate attributes and better image-text grounding, to improve the performance. 

\vspace{-0.2cm}
\subsection{Joint image \& language encoder}
\label{sec:encoder}
\vspace{-0.2cm}
The joint language and image encoder is illustrated in~\figref{fig:model}(b). The inputs to the joint encoder are word embeddings of a sentence (either $\mathbf{q}$ or $\mathbf{r}$) and associated object detections from $\rO$. The local image region defined by these bounding boxes is encoded via an image CNN to a $2048$ dimensional vector. This vector is projected to a $512$ dimensional embedding, using a fully connected \emph{downsample net}. The language and image embeddings are concatenated and transformed using a long-short term memory network (LSTM)~\cite{HochreiterNC1997}. 
Note that for  non-\emph{tag} words, \ie, words without an associated object detection, the object detection corresponding to the entire image is utilized.
The outputs of each unit of the LSTM are pooled together to obtain the final joint encoding of $\rQ$ (or $\rR$) and $\rO$. Note that the network components with a \mybox{black outline,} \ie, the downsample net and LSTM are the only components with trainable parameters. We design this so that no gradients need to be propagated back to the image CNN or to the BERT model, since both of them are parameter intensive, requiring significant training time and data. This choice facilitates the pre-computation of language and image features for faster training and inference. 

\vspace{-0.2cm}
\subsection{Improving visual representation \& image-text grounding}
\label{sec:improvements}
\vspace{-0.2cm}
\textbf{Attributes capturing visual features.} Almost all previous VCR baselines have used a CNN trained for ImageNet classification to extract visual features. Note that the class label $l_i$  for each bounding box is already available in the dataset and incorporated in the models (previous and ours) via BERT embeddings. 
We hypothesize that visual question answering and reasoning benefits from information about object characteristics and attributes. 
This intuition is illustrated in~\figref{fig:qual_results} where 
attributes add valuable information to help reason about the scene, such as `\emph{black} picture,' `\emph{gray} tie,' and `\emph{standing} man.'
To validate this hypothesis we deploy a pretrained attribute classifier which augments every detected bounding box $b_i$ with a set of attributes such as
colors, texture, size, and emotions. We show the attributes predicted by our model's image CNN in~\figref{fig:idea23}(a). 
\begin{figure}[t]
\begin{algorithm}[H]
\hspace*{-0.5cm}
\begin{minipage}{0.97\linewidth}
\algblockdefx[NAME]{WhileInParallel}{EndWhileInParallel}
      [1]{\textbf{while} #1 \textbf{in parallel do}}
          {\textbf{end}}
\begin{algorithmic}[1]

    \State Forward pass through image CNN to obtain object detections $\rhatO$
    \State $\cal{\hat{L}} \gets \mathtt{set}($all class labels in $\rhatO)$
    \For{$w \in \rW$ where $\rW \in \{\rQ, \rR\}$}
        \If{$w$ is tag}
            $w \gets \mathtt{remap}(w)$
        \EndIf
    \EndFor
    \State new\_tags $\gets \{ \}$
    \For{$w \in \rW$ where $\rW \in \{\rQ, \rR\}$}
        \If{($\mathtt{pos\_tag}(w|\rW) \in \{\mathtt{NN, NNS}\}$) and
            ($\mathtt{wsd\_synset}(w, \rW)$ has a noun)}
            \If{$w \in \cal{\hat{L}} $} \Comment{Direct match between word and detections}
                \State new\_detections $\gets$ detections in $\rhatO$ corresponding to $w$
                \State add ($w,$ new\_detections) to new\_tags
            \Else \Comment{Use word sense to match word and detections}
                \State max\_wup $\gets 0$
                \State word\_lemma $\gets \mathtt{lemma}(w)$
                \State word\_sense $\gets \mathtt{first\_synset}(\text{word\_lemma})$
                \For{$\hat{l} \in \cal{\hat{L}}$}
                    \If{$\mathtt{wup\_similarity}(\mathtt{first\_synset}(\hat{l}), \text{word\_sense}) > $ max\_wup}
                        \State max\_wup $\gets \mathtt{wup\_similarity}(\mathtt{
                        first\_synset
                        }(\hat{l}), \text{word\_sense})$
                        \State best\_label $\gets \hat{l}$ 
                    \EndIf
                \EndFor
            
                \If{max\_wup > $k$ 
                } 
                    \State new\_detections $\gets$ detections in $\rhatO$ corresponding to best\_label
                    
                    \State add ($w,$ new\_detections) to new\_tags
                    
                \EndIf
            \EndIf
        \EndIf
    \EndFor
\end{algorithmic}
\end{minipage}
  \caption{Finding \emph{new tags}}
  \label{alg:tagging}
\end{algorithm}
\vspace{-1.0cm}
\end{figure}
For this, we take advantage of work by~\citet{anderson2018bottom} as it incorporates attribute features to improve performance on language and vision tasks. Note that~\citet{zellers2019vcr}  evaluate the model proposed by~\citet{anderson2018bottom} with BERT embeddings to obtain  $39.6\%$ accuracy on the test set of the \qtoar~task. As detailed in~\secref{sec:quant}, with the same CNN and BERT embeddings, our network achieves $
50.5\%$. 
We achieve this by capturing recurrent information of LSTM modules via pooling and better scoring through an MLP. This is in contrast to~\citet{zellers2019vcr}, where the VQA 1000-way classification is removed and the response representation is scored using a dot product.

\textbf{\emph{New tags} for better text to image grounding.} 
Associating a word in the text with an object detection in the image, \ie, $o_i = (b_i, l_i)$ is what we commonly refer to as text-image grounding. Any word serving as a pointer to a detection is referred to as a \emph{tag} by~\citet{zellers2019vcr}.
Importantly, 
many nouns in the text (query or responses) aren't grounded with their appearance in the image. We explain possible reasons in \secref{sec:qual}. To overcome this shortcoming, we develop Algorithm~\ref{alg:tagging} to find new text-image groundings or \emph{new tags}. A qualitative example is illustrated in~\figref{fig:qual_results}. Nouns such as `cart' and `coats' weren't tagged by VCR, while our \shortmodel~model can tag them. 

Specifically, for text-image grounding we first find detections $\rhatO$ (in addition to VCR provided $\rO$) using the image CNN.  The set of unique class labels in $\rhatO$ is assigned to $\cal{\hat{L}}$. Both $\rQ$ and $\rR$ are modified such that all \emph{tags} (pointers to detections in the image) are remapped to natural language (class label of the detection). This is done via the \texttt{remap} function. We follow~\citet{zellers2019vcr} and associate a gender neutral name for the `person' class. For instance, ``How did [0,1] get here?'' in~\figref{fig:qual_results} is remapped to ``How did Adrian and Casey get here?''. This remapping is necessary for the next step of the part-of-speech (POS) tagging which operates only on natural language.

Next, the POS tagging function (\texttt{pos\_tag}) parses a sentence $\rW$ and assigns POS tags to each word $w$. For finding \textit{new tags}, we are only interested in words with the POS tag being either singular noun (NN) or plural noun (NNS). For these noun words, we check if a word $w$ directly matches a label in $\cal{\hat{L}}$. If such a direct match exists, we associate $w$ to the detections of the matching label. As shown in~\figref{fig:qual_results}(a), this direct matching associates the word \textbf{\uline{cart}} in the text (response 1 of the \qtoa\ subtask and response 4 of the \qator\ subtask) to the detection corresponding to label `cart' in the image, creating a \textit{new tag}. 

If there is no such direct match for $w$, we find matches based on word sense. This is motivated in~\figref{fig:qual_results}(b) where the word `coat' has no direct match to any image label in $\cal{\hat{L}}$. Rather there is a detection of `jacket' in the image. Notably, the word `coat' has multiple word senses, such as `an outer garment that has sleeves and covers the body from shoulder down' and `growth of hair or wool or fur covering the body of an animal.' Also, `jacket' has multiple word senses, two of which are `a short coat' and `the outer skin of a potato'. As can be seen, the first word senses of `coat' and `jacket' are similar and would help match `coat' to `jacket.' Having said that, the second word senses are different from common use and from each other. Hence, for words that do not directly match a label in $\cal{\hat{L}}$, choosing the appropriate word sense is necessary.
To this end, we adopt a simple approach, where we use the most frequently used word sense of $w$ and of labels in $\cal{\hat{L}}$. This is obtained using the first synset in Wordnet in NLTK~\cite{miller1995wordnet,loper2002nltk}.
Then, using the first synset of $w$ and labels in $\cal{\hat{L}}$, we find the best matching label `best\_label' corresponding to the highest Wu-Palmer similarity between synsets~\cite{wu1994verbs}. Additionally, we lemmatize $w$ before obtaining its first synset. 
 If the Wu-Palmer similarity between word $w$ and the `best\_label' is greater than a threshold $k$, we associate the word to the detections of `best\_label.' Overall this procedure leads to \textit{new tags} where text and label aren't the same but have the same meaning.
 We found $k=0.95$ was apt for our experiments. While inspecting, we found this algorithm missed to match the word `men' in the text to the detection label `man.' This is due to the `lemmatize' function provided by NLTK~\cite{loper2002nltk}. Consequently, we additionally allow \emph{new tags} corresponding to this `men-man' match. 

This algorithm permits to find \emph{new tags} in 
$7.1\%$ answers and $32.26\%$ rationales. A split over correct and incorrect responses is illustrated  in~\figref{fig:new-tag-stats}. These \emph{new tag} detections are used by our \emph{new tag} variant \textbf{\shortmodel}.
If there is more than one detection associated with
a \emph{new tag}, we average the visual features at the step before the LSTM in the joint encoder.

\begin{table}[t]
\centering
\small
\setlength{\tabcolsep}{5pt}
\begin{tabular}{lccccc}
\hline
\hline
                                                               & \multicolumn{1}{c}{\footnotesize \qtoa} & \multicolumn{1}{c}{\footnotesize \qator} & \multicolumn{1}{c}{\footnotesize \qtoar} & \multicolumn{2}{c}{\footnotesize Params (Mn)} 
                                                              \\ & (val)                                      & (val)                                       & (val)                                       &   {\scriptsize (total)}  &  {\scriptsize (trainable)}                
\\ \hline
R2C (\citet{zellers2019vcr})                                   & 63.8                                     & 67.2                                      & 43.1                                      & 35.3       & 26.8           
\\ \hline
\multicolumn{6}{c}{\emph{Improving R2C}}
\\ \hline
R2C + Det-BN                                                   & 64.49                                   & 67.02                                    & 43.61                                   & 35.3       & 26.8           \\
R2C + Det-BN + Freeze (\vcrpp)                                         & 65.30                                   &67.55                                    & 44.41                                    & 35.3       & 11.7           \\ 
\vcrpp~+ Resnet101                              & 67.55                             & 68.35                                     & 46.42                                     & 54.2       & 11.7           \\
\vcrpp~+ Resnet101 + Attributes                 & 68.53                                    & 70.86                                     & 48.64                                   & 54.0         & 11.5           \\ 
\hline
\multicolumn{6}{c}{\emph{Ours}}\\ 
\hline
Base                                          &  66.39                                    & 69.02                                    & 46.19                                     & 28.4       & 4.9            \\
Base + Resnet101                             & 67.50                                   & 69.75                                     & 47.51                                     & 47.4       & 4.9            \\
Base + Resnet101 + Attributes                & 69.51                                   & 71.57                                    & 50.08                                   & 47.2       & 4.7            \\
Base + Resnet101 + Attributes + New Tags (\textbf{\shortmodel}) & \textbf{69.89}                                    & \textbf{72.15}                                     & \textbf{50.62}                                     & 47.2       & 4.7\\ 
\hline
\hline\\
\end{tabular}
\vspace{-0.3cm}
    \caption{
    Comparison of our approach to the current state-of-the-art  R2C~\cite{zellers2019vcr} on the validation set. Legend: 
    \textbf{Det-BN}: Deterministic testing using train time batch normalization statistics.
    \textbf{Freeze}: Freeze all  parameters of the image CNN. 
    \textbf{ResNet101}: ResNet101 backbone as image CNN (default is ResNet50).
    \textbf{Attributes}: Attribute capturing visual features by using \cite{anderson2018bottom} (which has a ResNet101 backbone) as image CNN.
    \textbf{Base}: Our base model, as detailed in~\figref{fig:model}(b) and ~\secref{sec:overview}.
    \textbf{New Tags}: Augmenting object detection set with \emph{new tags} (as detailed in~\secref{sec:improvements}), \ie, grounding additional nouns in the text to the image.
    }
    \label{tab:final}
    \vspace{-0.6cm}
\end{table}
\textbf{Implementation details.} We  
defer specific details about training, implementation and design choices to the supplementary material. The code can be found at \url{https://github.com/deanplayerljx/tab-vcr}.

\section{Experiments}
\vspace{-0.2cm}
In this section, we first introduce the VCR dataset and describe metrics for evaluation. Afterward, we quantitatively compare our approach and improvements to the current state-of-the-art method~\cite{zellers2019vcr} and to top VQA models. We include a qualitative evaluation of \shortmodel~and an error analysis.

\vspace{-0.2cm}
\subsection{Dataset}
\label{sec:dataset}
\vspace{-0.2cm}
We train our models on the visual commonsense reasoning dataset~\cite{zellers2019vcr} which contains over 212k (train set), 26k (val set) and 25k (test set) questions
on over 110k unique movie scenes. The scenes were selected from LSMDC~\cite{rohrbach2017movie} and MovieClips,
after they passed an `interesting filter.'  
For each scene, workers were instructed to created `cognitive-level' questions. Workers answered these questions and gave a reasoning or \emph{rationale} for the answer. 
\begin{figure}[t]
    \begin{minipage}[b]{0.50\linewidth}
        \centering
        {\small
        \begin{tabular}{@{}c@{\hskip20pt}c@{\hskip15pt}c@{\hskip15pt}c@{}}
        \hline
        \hline
        \vspace{-2.5mm}\\
        Model
        & \qtoa             & \qator             & \qtoar             \\ \midrule
        Revisited~\cite{JabriECCV2016RevistVQA}
        & 57.5              & 63.5               & 36.8               \\
        BottomUp~\cite{anderson2018bottom}
        & 62.3              & 63.0               & 39.6               \\
        MLB~\cite{kim2017hadamard}
        & 61.8              & 65.4               & 40.6               \\
        MUTAN~\cite{BenyounesICCV2017Mutan}
        & 61.0              & 64.4               & 39.3               \\
        R2C~\cite{zellers2019vcr}
        & 65.1              & 67.3               & 44.0               \\
        \textbf{\shortmodel} (ours)
        & 
        \textbf{70.4}
        & 
        \textbf{71.7}
        & 
        \textbf{50.5}
        \\     
        \hline
        \hline
        \end{tabular}
        }
        \vspace{-0.2cm}
        \captionof{table}{\textbf{Evaluation on test set:} Accuracy on the three VCR tasks. 
        Comparison with top VQA models + BERT performance (source:~\cite{zellers2019vcr}). Our best model outperforms  R2C~\cite{zellers2019vcr} on the test set by a significant margin. 
        }
        \label{tab:test-set-results}
    \end{minipage}
    \hfill
    \begin{minipage}[b]{0.45\linewidth}
        \centering
        \includegraphics[width=\linewidth]{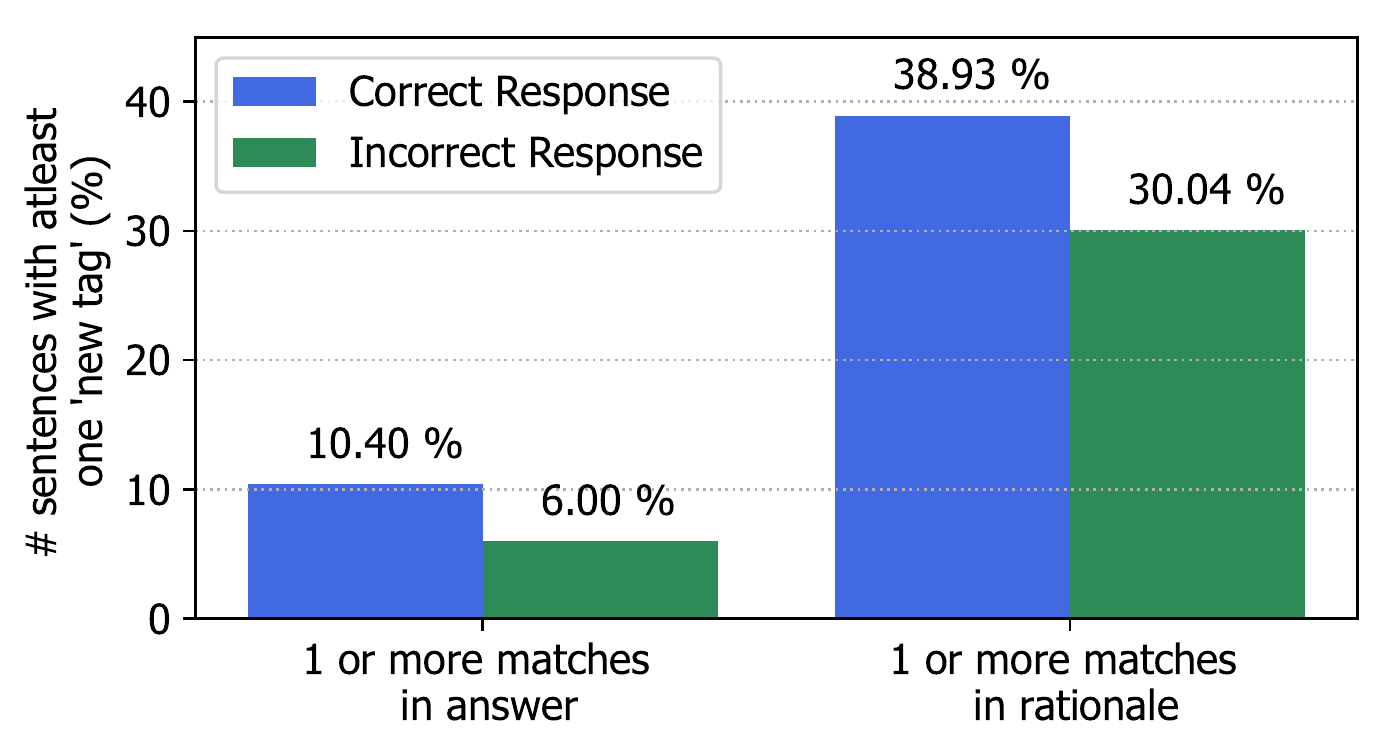}
        \vspace{-0.6cm}
        \captionof{figure}{\textbf{New tags:} Percentage of response sentences with a \emph{new tag}, \ie, a new grounding for noun and object detection. Correct responses  more likely  have new detections than incorrect ones. }
        \label{fig:new-tag-stats}
    \end{minipage}
    \vspace{-0.4cm}
\end{figure}

\vspace{-0.2cm}
\subsection{Metrics}
\label{sec:metrics}
\vspace{-0.2cm}
Models are evaluated with classification accuracy on the \qtoa, \qator~subtasks and the holistic \qtoar~task. For train and validation splits, the correct labels are available for development. To prevent overfitting, the test set labels were not released.
Since evaluation on the test set is a manual effort by~\citet{zellers2019vcr}, we provide numbers for our best performing model on the test set and illustrate results for the ablation study on the validation set. 

\vspace{-0.2cm}
\subsection{Quantitative evaluation}
\label{sec:quant}
\vspace{-0.2cm}

\tabref{tab:final} compares the performance of  variants of our approach to the current state-of-the-art R2C~\cite{zellers2019vcr}. While we report validation accuracy on both subtasks (\qtoa~and \qator) and the joint (\qtoar) task in~\tabref{tab:final}, in the following discussion we refer to percentages with reference to \qtoar. 

We make two modifications to improve R2C. The first is \texttt{Det-BN} where we calculate and use train time batch normalization~\cite{IoffeICML2015BatchNormalization} statistics. 
Second, we \texttt{freeze} all the weights of the image CNN in R2C, whereas~\citet{zellers2019vcr} keep the last block trainable. We provide a detailed study on \texttt{freeze} later. With these two minor 
changes, we obtain an improvement 
($1.31\%$) in performance and a significant reduction in trainable parameters (15Mn). We use the shorthand  \texttt{R2C++} to refer to this improved variant of R2C. 

\begin{figure}[t]
    \centering
    \begin{tabular}{@{\hskip-2pt}c@{\hskip-0pt}c@{\hskip-0pt}c}
    \includegraphics[width=0.33\textwidth]{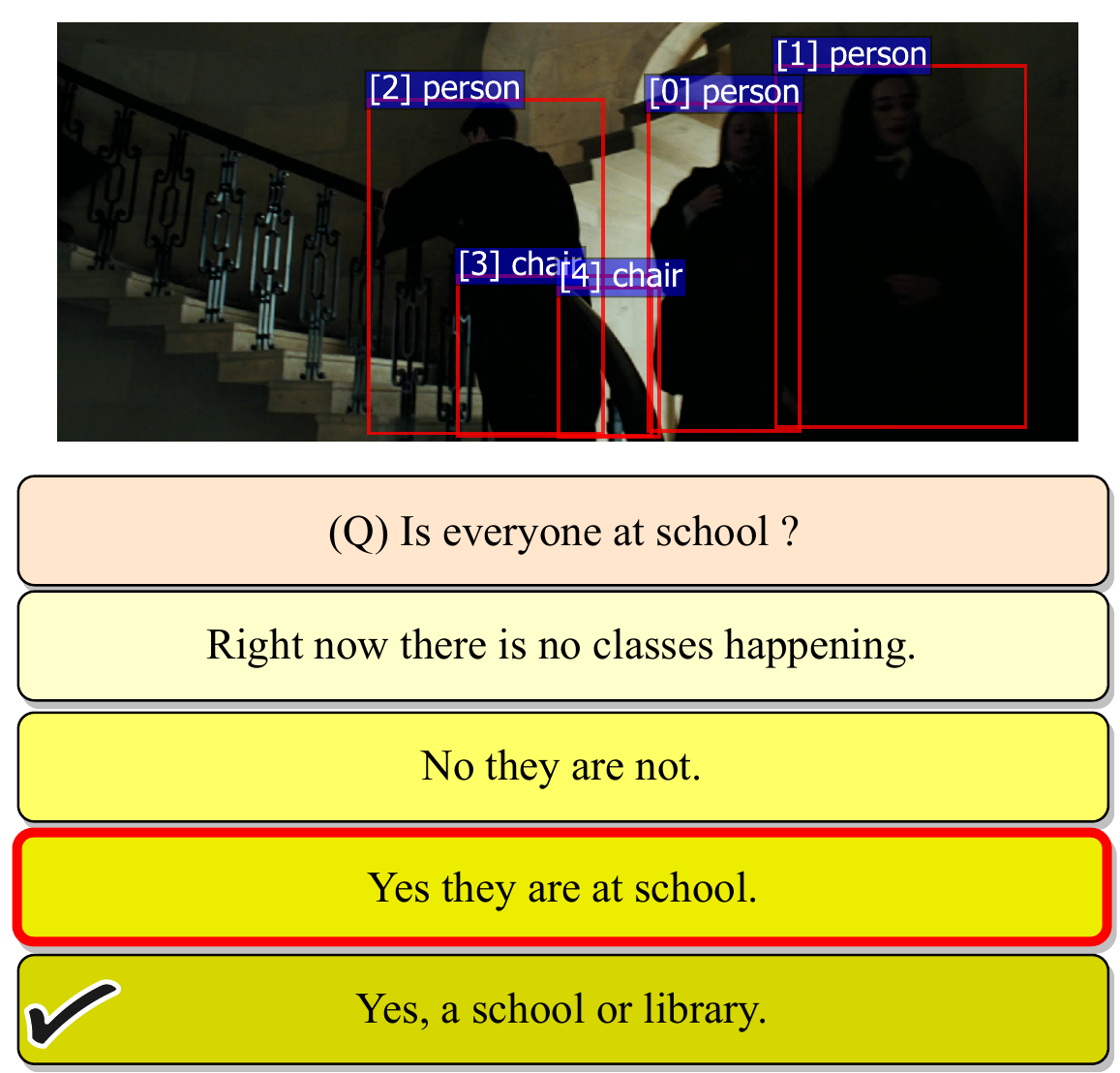} &
    \includegraphics[width=0.34\textwidth]{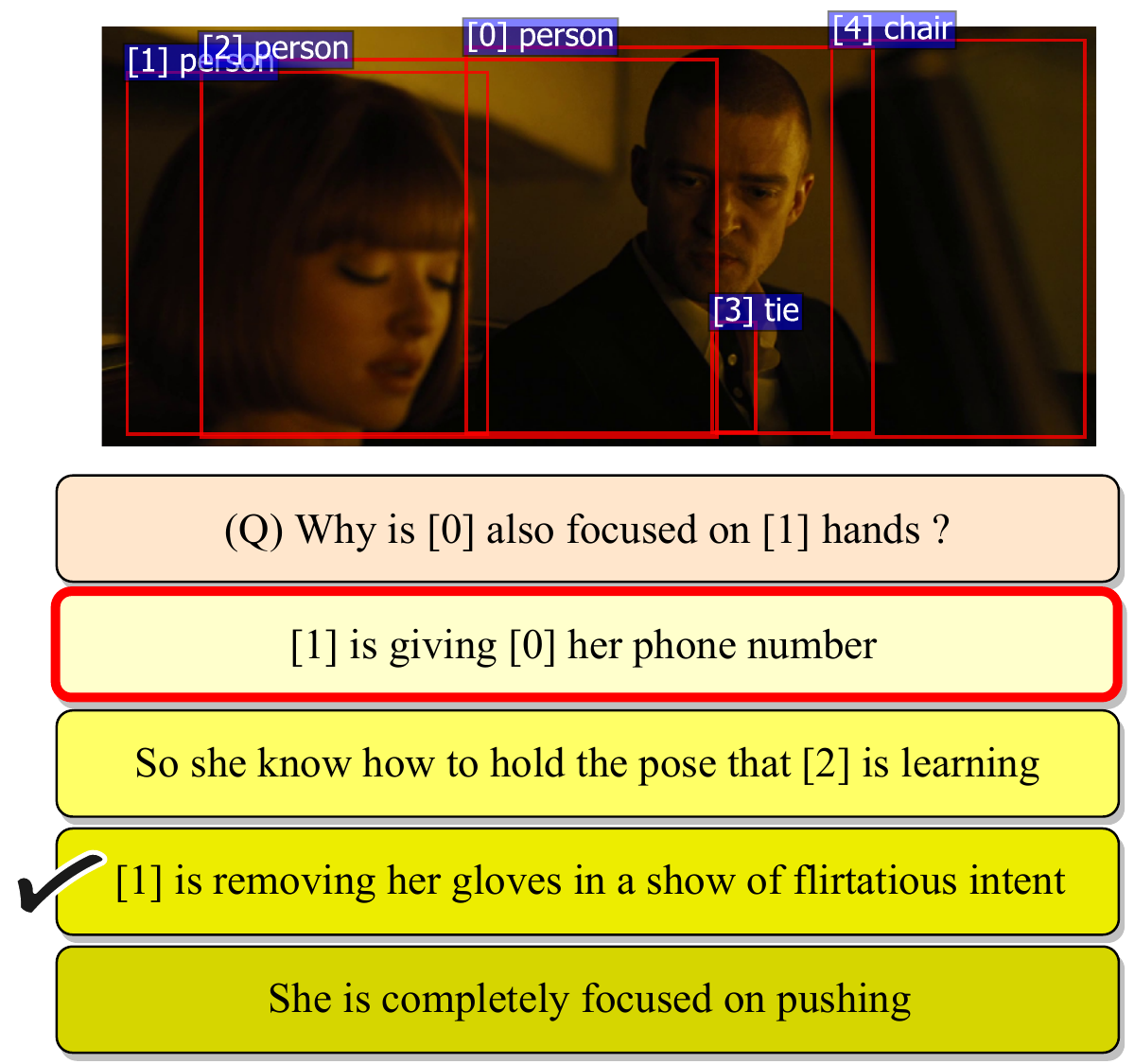} &
    \includegraphics[width=0.35\textwidth]{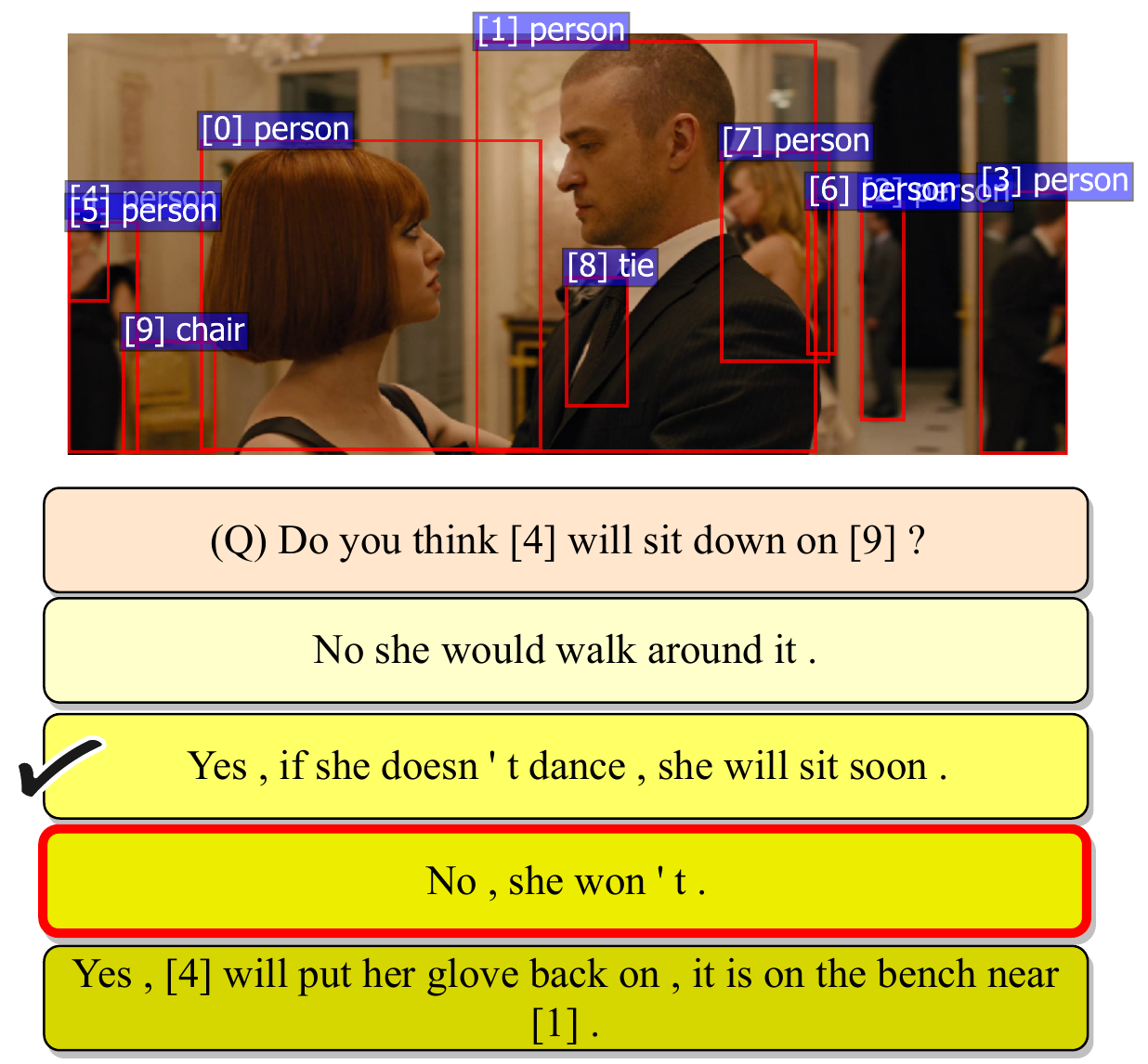}\\
    {\small (a) Similar responses}&
    {\small (b) Missing context}&
    {\small (c) Future ambiguity}\\
    \end{tabular}
    \vspace{-0.2cm}
    \caption{\textbf{Qualitative analysis of error modes:}  Responses with  similar meaning (left), lack of context (middle) or  ambiguity in future actions (right). Correct answers are marked with ticks and our model's incorrect prediction is outlined in red.}
    \label{fig:error-analysis}
    \vspace{-0.6cm}
\end{figure}

\begin{figure}[t]
    \begin{minipage}[t]{0.45\linewidth}
        \scriptsize
        \centering
        \begin{tabular}{ccccc}
        \hline
        \hline
        \vspace{-2.5mm}\\
        Encoder     &   \qtoa               & \qator            & \qtoar & Params\\
        \midrule
        Shared      &   69.89               & 72.15             & 50.62              & 4.7M                  \\
        Unshared    &   69.59               & 72.25             & 50.35              & 7.9M\\
        \hline
        \hline
        \end{tabular}
        \vspace{-0.2cm}
        \captionof{table}{Effect of shared \vs unshared parameters in the joint encoder $f(\hspace{2pt}\cdot\hspace{2pt}; \theta)$ of the \texttt{TAB-VCR} model.
        }
        \label{tab:sharing}
    \end{minipage}
    \hskip15pt
    \begin{minipage}[t]{0.49\linewidth}
        \scriptsize
        \centering
        \begin{tabular}{c|c|c|c}
            \hline
            \hline
            \vspace{-2.5mm}\\
            \multirow{2}{*}{\begin{tabular}{@{}c@{}}VCR subtask\end{tabular}}                  & \multicolumn{3}{c}{Avg. no. of {\em tags} in query+response}                     \\ \cline{2-4} 
                                        & (a) all                       & (b) correct & (c) errors\\ \midrule
            \qtoa& 2.673    & 2.719 & 2.566 \\
            \qator& 4.293    & 4.401 & 4.013 \\
            \hline
            \hline
        \end{tabular}
        \vspace{-0.2cm}
        \captionof{table}{Error analysis as a function of number of tags. Less image-text grounding increases \texttt{TAB-VCR} errors.}
        \label{tab:average_tags}
    \end{minipage}
\vspace{-5mm}
\end{figure}

\begin{figure}[t]
\begin{minipage}[t]{0.46\textwidth}
    \scriptsize
    \centering
    \begin{tabular}{@{}c@{\hskip3pt}l@{\hskip3pt}c@{\hskip2pt}|@{\hskip2pt}c@{\hskip3pt}c@{\hskip3pt}c@{\hskip4pt}c@{}}
    \hline
    \hline
    \vspace{-2.5mm}\\ 
    $\#$
    & \begin{tabular}{@{}c@{}}
    4th conv \\block
    \end{tabular} 
    & \begin{tabular}{@{}c@{}}
    Downsample\\net
    \end{tabular} 
    & \qtoa 
    & \qator 
    & \qtoar
    & \begin{tabular}{@{}c@{}}Trainable\\params (mn)\end{tabular} \\ \midrule
    1& \tune                    & \present        & 64.57             & 68.86              & 44.60             & 19.9                 \\
    2& \tune\hspace{1mm} ($1/2$)                   & \present        &     64.26        &    68.14         &    44.08       & 19.9                 \\
    3& \tune\hspace{1mm} ($1/4$)                    & \present        & 63.11             & 67.73              & 42.87             & 19.9                 \\
    4& \tune\hspace{1mm} ($1/8$)                    & \present        & 63.51             &      67.49         &       43.21             & 19.9                 \\
    5& \freeze                   & \present        & 66.47             & 69.22              & 46.45             & 4.9                  \\
    6& \freeze                   & \absent    & 65.30             & 69.09              & 45.57             & 7.0                  \\
    \hline
    \hline
    \end{tabular}
    \vspace{-0.2cm}
    \captionof{table}{Ablation for \texttt{base} model: \tune\hspace{0.5mm}: Finetuning and \freeze\hspace{0.5mm}: Freezing weights of the fourth conv block in ResNet101 image CNN. Presence and absence of downsample net (to project image representation from 2048 to 512) is denoted by \present\hspace{1mm}and\hspace{1mm}\absent.}
    \label{tab:finetuning}
\end{minipage}
\hskip25pt
\begin{minipage}[t]{0.44\textwidth}
    \scriptsize
    \centering
    \begin{tabular}{@{\hskip4pt}c@{\hskip4pt}c@{\hskip4pt}c@{\hskip4pt}|@{\hskip4pt}c@{\hskip4pt}c@{}}
        \hline
        \hline
        \vspace{-2.5mm}\\
        \begin{tabular}{@{}c@{}}
            Ques. type
        \end{tabular}
        &  Matching patterns                 
        & Counts 
        & \begin{tabular}{@{}c@{}}
            \qtoa
        \end{tabular}
        & \begin{tabular}{@{}c@{}}
            \qator
        \end{tabular}
        \\ \midrule
        what    &  what                     & 10688 & 72.30 & 72.74\\
        why     &  why                      & 9395  & 65.14& 73.02\\
        isn't     &  is, are, was, were, isn't  & 1768  & 75.17 & 67.70\\
        where   &  where                    & 1546  & 73.54 & 73.09\\
        how     &  how                      & 1350  & 60.67 & 69.19\\
        do      &  do, did, does            & 655   & 72.82 & 65.80\\
        who     &  who, whom, whose         & 556   & 86.69 & 69.78\\
        will    &  will, would, wouldn't      & 307   & 74.92 & 73.29\\
        \hline
        \hline
    \end{tabular}
    \vspace{-0.1cm}
    \captionof{table}{Accuracy by question type (with at least 100 counts) of \texttt{TAB-VCR} model. \emph{Why} \& \emph{how} questions are most challenging for the \qtoa\ subtask.
    }
    \label{tab:question_type}
\end{minipage}
\vspace{-2mm}
\end{figure}

Our \texttt{base} model (described in~\secref{sec:approach}) which includes (\texttt{Det-BN}) and \texttt{Freeze} improvements,
improves over \texttt{R2C++} by 
$1.78\%$, while being conceptually simple,  having half the number of trainable parameters. 
By using a more expressive ResNet as image CNN model (\texttt{Base + Resnet101}), we obtain another 
$1.32\%$ improvement. We obtain another big increase of 
$2.57\%$ by leveraging attributes capturing visual features (\texttt{Base + Resnet101 + Attributes}). Our best performing variant incorporates \emph{new tags} during training and inference (\shortmodel) with a final 
$50.62\%$ on the validation set. We ablate \texttt{R2C++} with  \texttt{ResNet101} and \texttt{Attributes} modifications, which leads to better performance too. This suggests our improvements aren't confined to our particular net.
Additionally, we share the encoder for query and responses. We empirically studied the effect of sharing encoder parameters and found no significant difference (\tabref{tab:sharing}) when using separate weights, which comes at the cost of 3.2M extra trainable parameters. Note that~\citet{zellers2019vcr} also share the encoder for query and response processing. Hence, our design choice makes the comparison fair.

In \tabref{tab:test-set-results} we show results evaluating the performance of \shortmodel~on the private test set, set aside by~\citet{zellers2019vcr}. We obtain a 
5.3\%, 4.4\% and 6.5\% absolute improvement over R2C on the test set. We perform much better than top VQA models which were adapted for VCR in~\cite{zellers2019vcr}. 
Models evaluated on the test set are posted on the leaderboard\footnote{\url{visualcommonsense.com/leaderboard}}. We appear as `TAB-VCR' and outperform prior peer-reviewed work. At the time of writing ($23^\text{rd}$ May 2019)~\shortmodel~ranked second in the single model category.
After submission of this work other reports addressing VCR have been released. At the time of submitting this camera-ready ($27^\text{th}$ Oct 2019),~\shortmodel~ranked seventh among single models  on the leaderboard.
Based on the available reports~\cite{li2019visualbert, su2019vl,alberti2019fusion,li2019unicoder,lu2019vilbert,chen2019uniter}, most of these seven methods capture the idea of re-training BERT with extra information from Conceptual Captions~\cite{sharma2018conceptual}. This, in essence, is orthogonal to our \textit{new tags} and attributes approach to build simple and effective baselines with significantly fewer parameters.

\figref{fig:new-tag-stats} illustrates the effectiveness of our \emph{new tag} detection, where $10.4\%$ correct answers had at least one \emph{new tag} detected. With $38.93\%$, the number is even higher for correct rationales. This is intuitive as humans refer to more objects while reasoning about an answer than the answer itself.

\noindent\textbf{Finetuning \vs freezing last conv block. }In \tabref{tab:finetuning} we study the effect of finetuning the last conv block of ResNet101 and the downsample net.~\citet{zellers2019vcr} use row \#1.  We assess  lower learning rates -- 0.5x, 0.25x, and 0.125x (\#2 to \#4). We chose to freeze the conv block (\#5) to reduce trainable parameters by 15M, with slight improvement in performance. By comparing \#5 and \#6, we find  the presence of downsample net to reduce the model size and improve performance. After conducting this ablation study for the \texttt{base} model's architecture design, we updated the python dependency packages. This update lead to a slight difference in the accuracy of \#5 in \tabref{tab:finetuning} (before the update) and the final accuracy reported in~\tabref{tab:final} (after the update). 
However, the versions of python dependencies are consistent across all variants listed in~\tabref{tab:finetuning}. 

\vspace{-0.2cm}
\subsection{Qualitative evaluation and error analysis}
\label{sec:qual}
\vspace{-0.2cm}
We illustrate the qualitative results in~\figref{fig:qual_results}. We separate the image input to our model into three parts, for easy visualization. We show VCR detections \& labels, attribute prediction of our image CNN and \emph{new tags} in the left, middle and right images.
Note how our model can ground important words. 
For instance, for the example shown in~\figref{fig:qual_results}(a), 
the correct answer and rationale prediction is based on the cart in the image, which we ground. The word 
\textbf{\uline{`cart'}} wasn't  grounded in the original VCR dataset. Similarly, grounding the word \textbf{\uline{coats}} helps to answer and reason about the example in~\figref{fig:qual_results}(b).

\noindent\textbf{Explanation for missed tags.} As discussed in~\secref{sec:improvements}, the VCR dataset contains various nouns that aren't tagged such as `eye,' 
`coats' and `cart' as highlighted in~\figref{fig:idea23} and \figref{fig:qual_results}. 
This could be accounted to the methodology adopted for collecting the VCR dataset.~\citet{zellers2019vcr} instructed workers to provide questions, answers, and rationales by using natural language and object detections $\rO$ 
(COCO~\cite{lin2014microsoft} objects). We found that workers used natural language even if the corresponding object detection was available. Additionally, for some data points, we found objects mentioned in the text without a valid object detection in $\rO$. 
This may be because the detector used by~\citet{zellers2019vcr} is trained on COCO~\cite{lin2014microsoft}, which has only 80 classes.

\noindent\textbf{Error modes.} We also qualitatively study \shortmodel's shortcomings by analyzing error modes, as illustrated in~\figref{fig:error-analysis}. 
The correct answer is marked with a tick while our prediction is outlined in red. Examples include options with overlapping meaning (\figref{fig:error-analysis}(a)). Both the third and the fourth answers have similar meaning which could be accounted for the fact that~\citet{zellers2019vcr} automatically curated  competing incorrect responses via adversarial matching. Our method misses the `correct' answer.  Another error mode (\figref{fig:error-analysis}(b)) is due to objects which aren't present in the image, like the ``gloves in a show of flirtatious intent.'' This could be accounted to the fact that crowd workers were shown context from the video in addition to the image (video caption), which isn't available in the dataset. Also, as highlighted in \figref{fig:error-analysis}(c), scenes often  offer an ambiguous future, and our model gets some of these cases incorrect.

\noindent\textbf{Error and grounding.}
In \tabref{tab:average_tags}, we provide the average number of tags in the query+response for both subtasks. We state this value for the following subsets: (a) all datapoints, (b) datapoints where~\shortmodel\ was correct, and (c) datapoints where~\shortmodel\ made errors. Based on this, we infer that our model performs better on datapoints with more tags, \ie, richer association of image and text.

\noindent\textbf{Error and question types.} In \tabref{tab:question_type} we show the accuracy of the TAB-VCR model based on question type  defined by the corresponding matching patterns. Our model is more error-prone on \emph{why} and \emph{how} questions on the \qtoa\ subtask, which usually require more complex reasoning.
\vspace{-0.2cm}
\section{Conclusion}
\vspace{-0.2cm}
We develop a simple yet effective baseline for visual commonsense reasoning. The proposed approach leverages additional object detections to better ground noun-phrases and assigns attributes to current and newly found object groundings. Without an intricate and meticulously designed attention model, we show that the proposed approach outperforms state-of-the-art, despite significantly fewer trainable parameters. We think this simple yet  effective baseline and the new noun-phrase grounding can provide the basis for further development of visual commonsense models.

\clearpage
\section*{Acknowledgements}
This work is supported in part by NSF under Grant No.\ 1718221 and MRI \#1725729, UIUC, Samsung, 3M, Cisco Systems Inc.\ (Gift Award CG 1377144) and Adobe. We thank NVIDIA for providing GPUs used for this work and Cisco for access to the Arcetri cluster. The authors thank Prof. Svetlana Lazebnik for insightful discussions and Rowan Zellers for releasing and helping us navigate the VCR dataset \& evaluation.

{\small
\bibliographystyle{abbrvnat}
\bibliography{tbone}
}
\clearpage

\section{Supplementary Material for TAB-VCR: Tags and Attributes based Visual Commonsense Reasoning Baselines}
We structure the supplementary into two subsections.

\begin{enumerate}
    \compresslist
    \item Details about implementation and training routine, including hyperparamters and design choices.
    \item Additional qualitative results including error modes
    \item Log of version changes
\end{enumerate}
\subsection{Implementation and training details}
\begin{figure}[h]
    \centering
    \begin{tabular}{@{\hskip-4pt}cc}
        \includegraphics[width=0.5\linewidth]{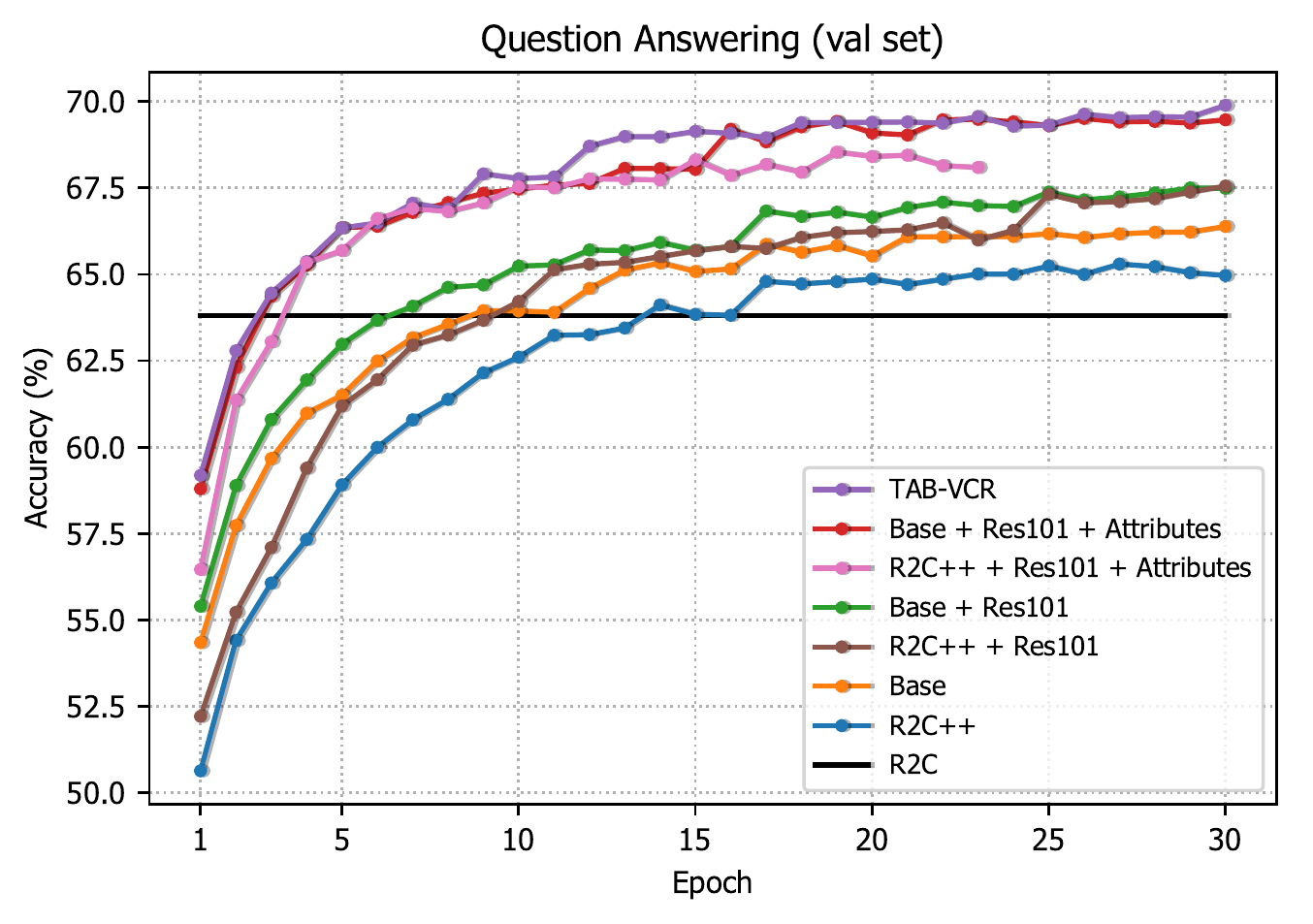}&
        \includegraphics[width=0.5\linewidth]{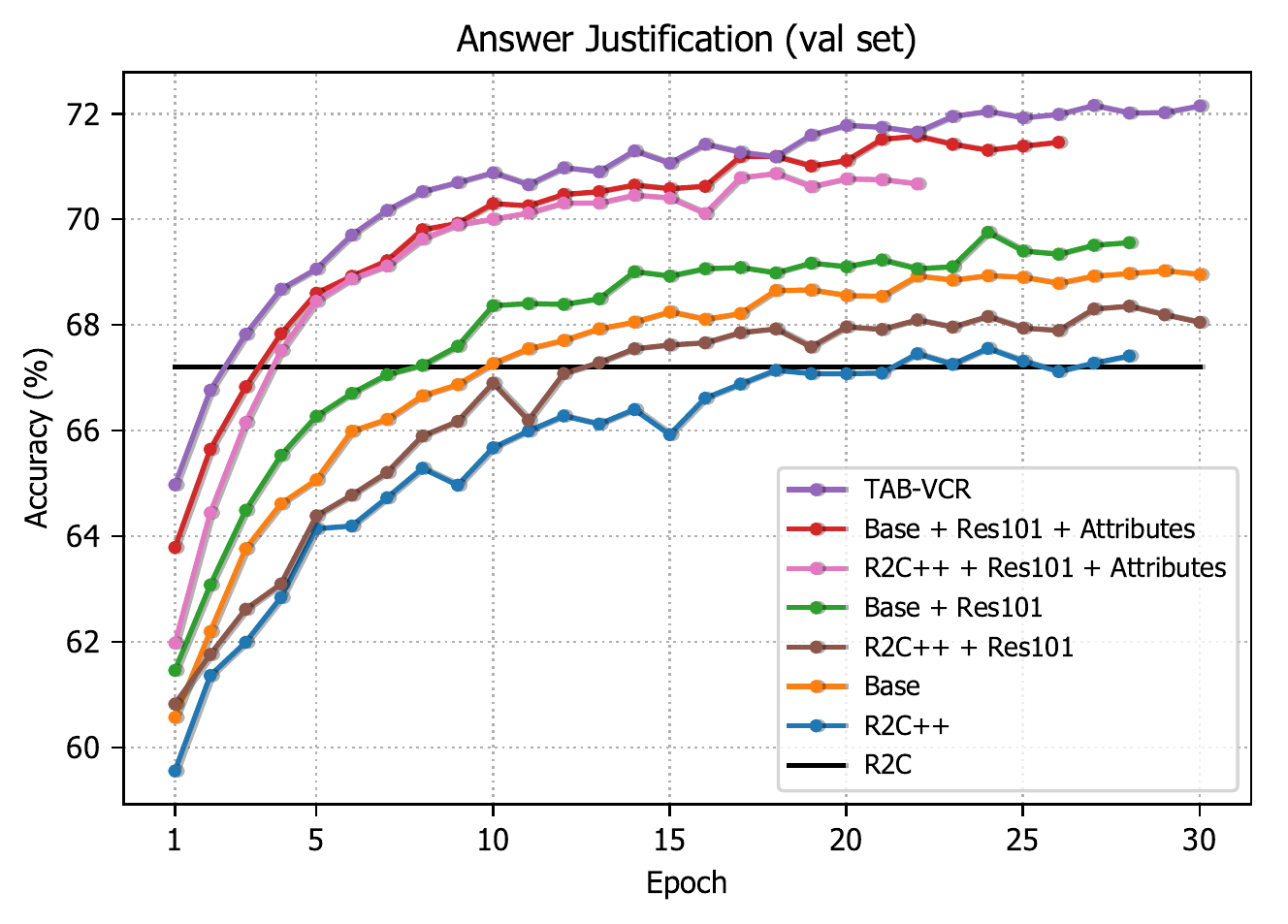}\\
    \end{tabular}
    \caption{\textbf{Accuracy on validation set.} Performance for \qtoa~(left) and \qator~(right) tasks. } 
    \label{fig:val-plots}
\end{figure}
As explained in~\secref{sec:overview}, our approach is composed of three components. Here, we provide implementation details for each:  (1) BERT: Operates over query and response under consideration. The features of the penultimate layer are extracted for each word. \citet{zellers2019vcr} release these embeddings with the VCR dataset and we use them as is. (2) Joint encoder: As detailed in~\secref{sec:quant},  we assess different variants over the baseline model using two CNN models. The output dimension of each  is 2048. The downsample net is a single fully connected layer with input dimension of 2048 (from the image CNN) and an output dimension of 512. We use a bidirectional LSTM with a hidden state dimension of $2\cdot 256=512$. The outputs of which are average pooled. (3) MLP: Our MLP is much slimmer than the  one from the  R2C model. The pooled query and response representations are concatenated to give a  $512+512=1024$ dimensional input. The MLP has a $512$ dimensional hidden layer and a final output (score) of dimension 1.
The threshold for Wu Palmer similarity $k$ is set to $0.95$.

We used the cross-entropy loss function for end-to-end training, Adam optimizer with learning rate  $2\mathrm{e}{-4}$, and LR scheduler that reduce the learning rate by half after two consecutive epochs without improvement. We train our model for 30 epochs. We also employ early stopping, \ie, we stop training after 4 consecutive epochs without validation set improvement. \figref{fig:val-plots} shows validation accuracy for both the subtasks of VCR over the training epochs. We observe the proposed approach to very quickly exceed the results reported by previous state-of-the-art (marked via a solid horizontal black line).
\subsection{Additional qualitative results}
\begin{figure}[h]
\vspace{-1cm}
    \centering
    \includegraphics[width=\linewidth]{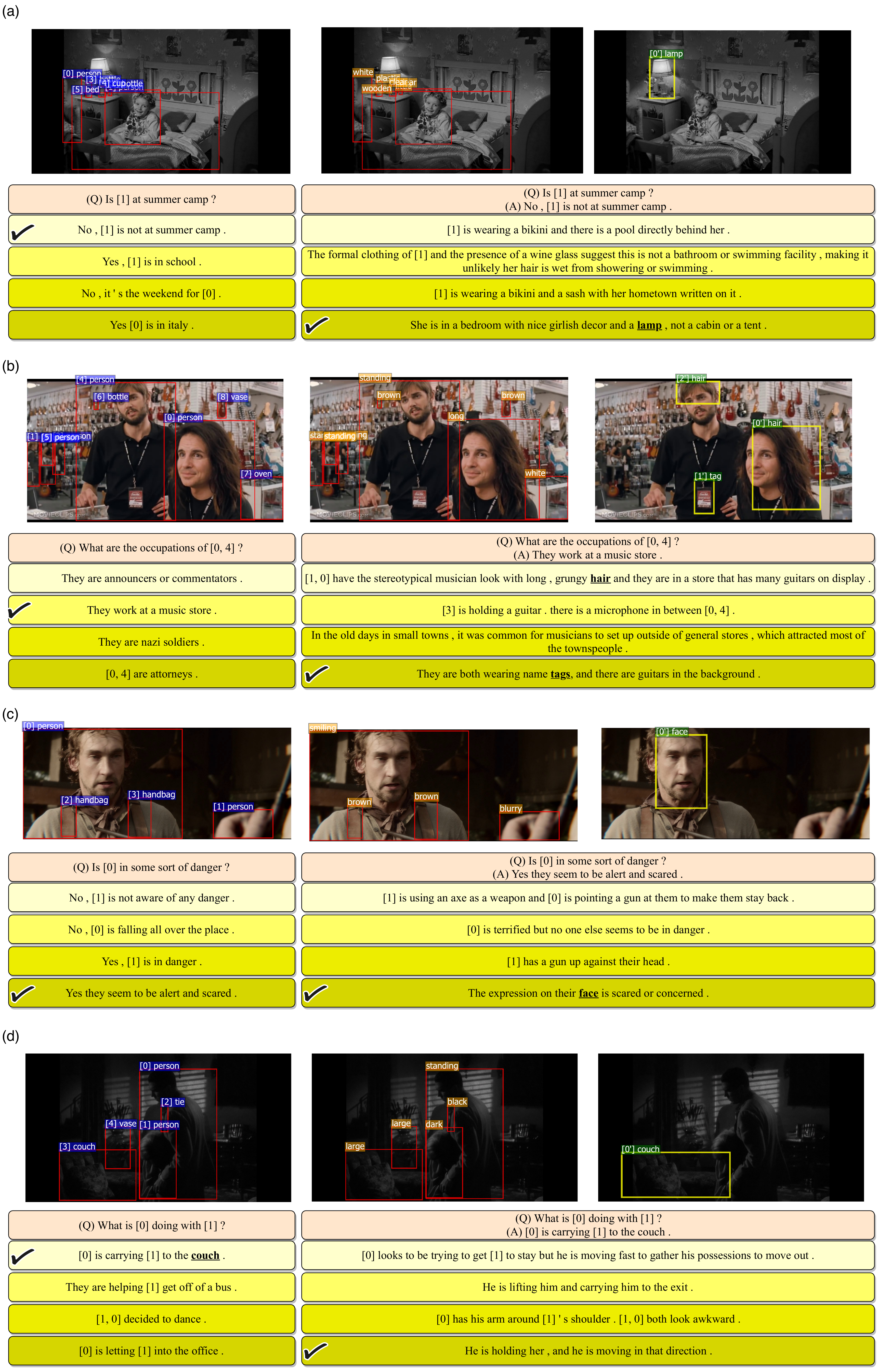}
    \caption{\textbf{Qualitative results.} More examples of the proposed \textbf{\shortmodel} model, which incorporates attributes and augments image-text grounding. The image on the left shows the object detections provided by VCR. The image in the middle shows the attributes predicted by our model and thereby captured in visual features. The image on the right shows \emph{new tags} detected by our proposed method. Below the images are the question answering and answer justification subtasks. The \emph{new tags} are highlighted in \textbf{\uline{bold}}.}
    \label{fig:qual_good_supp}
\end{figure}

Examples of \shortmodel~performance on the VCR dataset are included in~\figref{fig:qual_good_supp}. They supplement  the qualitative evaluation in the main paper (\secref{sec:qual} \&~\figref{fig:qual_results}). Our model correctly predicts for each of these examples. Note how our model can ground important words. These are highlighted in \textbf{\uline{bold}}. For instance, for  \figref{fig:qual_good_supp}(a), the correct rationale prediction is based on the expression of the \textbf{\uline{lamp}}, which we ground. The lamp wasn't grounded in the original VCR dataset. Similarly grounding the \textbf{\uline{tag}}, and \textbf{\uline{face}} helps answer and reason for the image in~\figref{fig:qual_good_supp}(b) and~\figref{fig:qual_good_supp}(c). As illustrated via the \textbf{\uline{couch}} in~\figref{fig:qual_good_supp}(d), it is interesting that the same noun is present in detections yet not grounded to words in the VCR dataset. This could be accounted to the data collection methodology, as explained in~\secref{sec:qual} (`explanation of missed tags') of the main paper.

\begin{figure}[h]
    \centering
    \includegraphics[width=\linewidth]{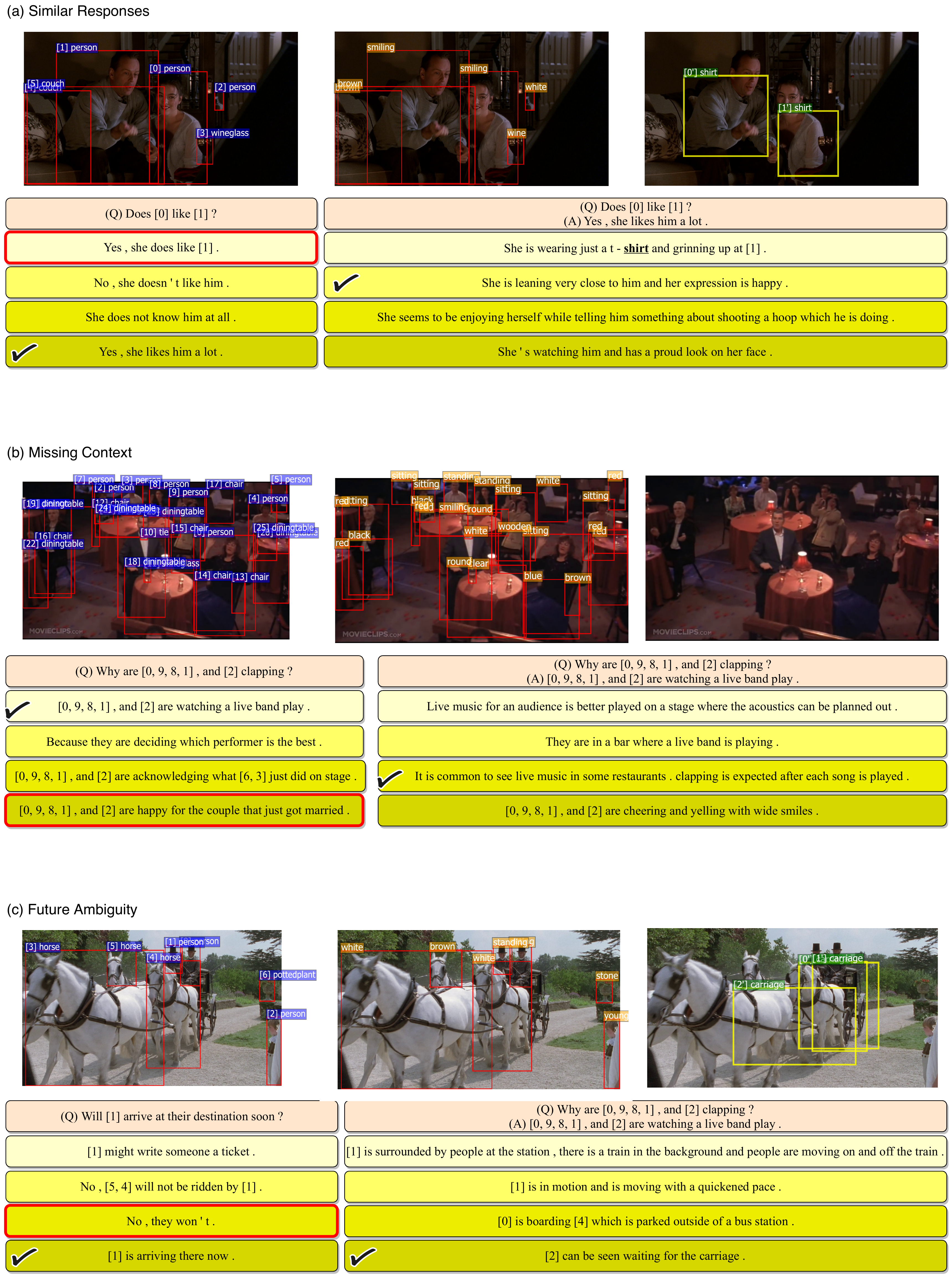}
    \caption{\textbf{Qualitative analysis of error modes.}  Responses with (a) similar meaning, (b) lack of context and (c) ambiguity in future actions. Correct answers are marked with ticks and our models incorrect prediction is outlined in red.}
    \label{fig:qual_error_supp}
\end{figure}

In~\figref{fig:qual_error_supp}(a), we provide additional examples to supplement the discussion of error modes in the main paper (\secref{sec:qual} \&~\figref{fig:error-analysis}).~\shortmodel~gets the question answering subtask (left) incorrect, which we detail next. Once the model knows the correct answer it can correctly reason about it, as  evidenced by being correct on the answer justification subtask (right).  In~\figref{fig:qual_error_supp}(a) both the responses `Yes, she does like [1]' and `Yes, she likes him a lot' are very similar, and our model misses the `correct' response. Since the VCR dataset is composed by an automated adversarial matching, these options could end up being very overlapping and cause these errors.  In~\figref{fig:qual_error_supp}(b) it is difficult to infer that the the audience are watching a live band play. This could be due to the missing context as video captions aren't available to our models, but were available to  workers during dataset collection. In~\figref{fig:qual_error_supp}(c)  multiple stories could follow the current observation, and \shortmodel~makes errors in  examples with  ambiguity regarding the future.

\subsection{Change Log} \textbf{v1.} First Version. \textbf{v2.} NeurIPS 2019 camera ready version with edits to rectify class labels in ~\figref{fig:idea23}, ~\figref{fig:qual_results}, and ~\figref{fig:qual_error_supp}.

\end{document}